\documentclass{bmvc2k}
\usepackage{comment}
\usepackage{todonotes}
\usepackage{enumerate}
\usepackage{soul}
\usepackage{graphicx}
\usepackage{mathtools}
\usepackage{amsfonts}
\usepackage{url}

\title{Detection-Fusion for Knowledge Graph Extraction from Videos}

\addauthor{Taniya Das}{t.das@tue.nl}{1}
\addauthor{Louis Mahon}{lmahon@ed.ac.uk}{1,2}
\addauthor{Thomas Lukasiewicz}{thomas.lukasiewicz@tuwien.ac.at}{1,3}

\addinstitution{
University of Oxford,\\
Oxford, UK}
\addinstitution{
University of Edinburgh,\\
Edinburgh, UK}
\addinstitution{
Vienna University of Technology,\\
Vienna, Austria}

\runninghead{Das, Lukasiewicz, Mahon}{Detection-Fusion for KG Extraction from Videos}

\begin{document}

\maketitle

\begin{abstract}
One of the challenging tasks in the field of video understanding is extracting semantic content from video inputs. Most existing systems use language models to describe videos in natural language sentences, but this has several major shortcomings. Such systems can rely too heavily on the language model component and base their output on statistical regularities in natural language text rather than on the visual contents of the video. Additionally, natural language annotations cannot be readily processed by a computer, are difficult to evaluate with performance metrics and cannot be easily translated into a different natural language. In this paper, we propose a method to annotate videos with knowledge graphs, and so avoid these problems. Specifically, we propose a deep-learning-based model for this task that first predicts pairs of individuals and then the relations between them. Additionally, we propose an extension of our model for the inclusion of background knowledge in the construction of knowledge graphs. %To the best of our knowledge, no existing work has explored this task of background knowledge inclusion. 
\end{abstract}

%-------------------------------------------------------------------------
\section{Introduction}
\label{sec:introduction}
Visual understanding has been a central question in AI since the inception of the field. However, it is not obvious how to quantify whether a machine can understand what it sees. One simple way is classification, and indeed, much of the computer vision research over the last ten years has centered around ImageNet. Object classification performance is very easy to measure, but it only conveys a coarse description of the image 
%by identifying the presence of an object or objects. 
and misses further information about the properties and relations of the present objects. Another approach is to generate a natural language sentence describing the visual contents. This escapes the limitation of classification and is capable of expressing all the complexity that natural language can express. 

However, using natural language comes with a number of disadvantages. It means the model not only has to learn to understand the contents of the video but also how to express this content in natural language, which is a significant additional requirement. Even in humans, understanding is quite a separate problem from articulation in language, as evidenced by patients with damage to Broca's area in the brain, which show normal understanding of visual and even linguistic information \citep{berndt1980redefinition}, but struggle to articulate this understanding in language \citep{thompson2012introduction}. Additionally, all the extra structure learned by the language generation component can obscure the performance of the understanding component. For example, if an image of a dog running in a park was correctly captioned as ``a dog is running in a park'', then we cannot conclude that model correctly identified a park and the action of running in the image. Instead, it may have identified the dog, and then the language model simply completed the most likely sentence that begins with ``a dog...''. Another problem with natural language annotations is that they are difficult to evaluate. The complex syntactic-semantic structure of a sentence means that we cannot simply count which words the model predicted correctly, but instead must use a bespoke metric such as BLEU \citep{papineni2002bleu}, METEOR \citep{banerjee2005meteor}, or LEPOR \citep{han2012lepor}. Recognizing the imperfection of each of these, results for natural-language annotation models typically report scores on multiple metrics, none of which have a simple and intuitive interpretation. A third disadvantage of requiring the model to produce a natural language annotation is that it commits it to that particular natural language. A model trained to produce English captions cannot, then, be used to produce Turkish captions. Not only has the model learned a different vocabulary, but a different grammar too, e.g. nominative, SVO, and analytic for English, vs ergative, SOV, and agglutinative for Turkish. Thus, as well as leading to unnecessary extra work, producing annotations in natural language hinders the generalization of the model. 

We instead choose to annotate videos using structured annotations in the form of knowledge graphs, which avoids all of the above drawbacks. Knowledge graphs, which are equivalent to sets of logical facts, can be evaluated with accuracy and similar metrics such as F1-score, do not require learning language syntax, and can be translated between natural languages by translating one term at a time (we can avoid word-sense disambiguation by relating to words at the sense-level). The first stage of our proposed model is to separately detect the individuals and predicates that are present in the input video, and then fuse the detected components together into a knowledge graph. Individuals are represented as learnable vectors; predicates, as multi-layer perceptrons (MLPs). We predict a fact as true if the output of the MLP, when input with the vector, is greater than a threshold. Each fact contains a predicate and the corresponding arguments: $<subject, predicate, object>$ (for binary facts) and $<subject, predicate>$ (for unary facts).

Our proposed model significantly outperforms existing works at the important task of annotating videos with knowledge graphs. We also explore the inclusion of background knowledge in the construction of the knowledge graphs. To the best of our knowledge, no existing work has explored this.

To summarize our contributions,
\begin{itemize}
    \item We propose a new deep-learning model to annotate videos with knowledge graphs. This is a superior approach to the more common one of annotating with natural language because it avoids learning unnecessary language syntax, is easier to evaluate, and can be translated easily to different natural languages. 
    \item We show experimentally that our proposed model significantly outperforms existing works that aim to annotate videos with knowledge graphs. 
    \item We explore the inclusion of background knowledge in the extraction of the knowledge graphs, the first work to do so.
    \item We present extensive ablation studies showing the contribution of each component of our model, and showing a trade-off between increased run-time and increased accuracy by varying number of individuals and predicates evaluated in the second stage.
\end{itemize}

The rest of this paper is organized as follows: Section~\ref{sec:related-work} gives an overview of related work, Section~\ref{sec:method} describes our proposed method, Section~\ref{sec:results} presents our experimental evaluation, and Section~\ref{sec:conclusion} summarizes our work.

\section{Related Work} \label{sec:related-work}
In \citeyear{johnson2015image}, \citeauthor{johnson2015image} advocated for the annotation of images using scene graphs, which describe the semantic and spatial properties and relations between objects in the image. Many following works addressed the task of forming structured annotations of still images, and it is now a reasonably well-established task in computer vision \cite{lu2016visual,xu2017scene,yang2018graph,chen2019knowledge}.
Scene-graph construction can be extended from images to videos. The resulting task, video scene-graph construction, is similar to our task of knowledge graph extraction. Both express the individuals, properties and relations in the input video. The crucial difference is that models which apply scene-graph extraction methods that were designed for images, have to process each frame separately, and then attempt to merge the graphs for each frame into one graph for the entire video. Various complicated methods have been proposed to this end \citep{teng2021target,tsai2019video,shang2017video}. The method of  \cite{liu2020beyond} is slightly different in that it first combines the objects and relations across frames, and then uses these to produce a single set of logical facts. However, it still differs significantly from our work in that we do not use tubes at all but rather have a single classifier for all frames, and then use a single learnable vector for all instances of the same object, which allows sharing of representation power across different videos. \cite{vasile2018learning} propose to generate logical facts as strings, using a language model output head, but this falsely interprets the knowledge graph as ordered.

The most similar existing work to ours is that of \cite{mahon2020knowledge}, which predicts the individuals present in a video, and then runs an MLP for each predicate on each individual and pair of individuals to form a single knowledge graph for the entire video. The key difference in our method is that we also predict the predicates, and then use a novel method of combining the predictions for subject, predicate and object. We also differ in the inclusion of background knowledge, as shown in Section \ref{sec:related-work}, though the reason we significantly outperform \cite{mahon2020knowledge} is mostly the architecture change, rather than the background knowledge.

As well as the general goal of providing a compact, largely language-agnostic description of video contents, some works have employed structured annotations for more specific purposes. \cite{li2022embodied} generate scene graphs from videos in the context of robot movement. That is, the robot moves around in the environment while taking video that the annotation is made of.
\cite{ross2018grounding} uses structured annotations, in particular lambda expressions, which are equivalent to sets of facts from first-order logic. They take a dataset of videos and paired sentences, and then use their generated lambda expression, to train a semantic parser without supervision on the natural language sentence.

\section{Method} \label{sec:method}
\subsection{Main Model}
\label{sec:main_model}

\begin{figure*}
    \centering
    \includegraphics[width=\textwidth]{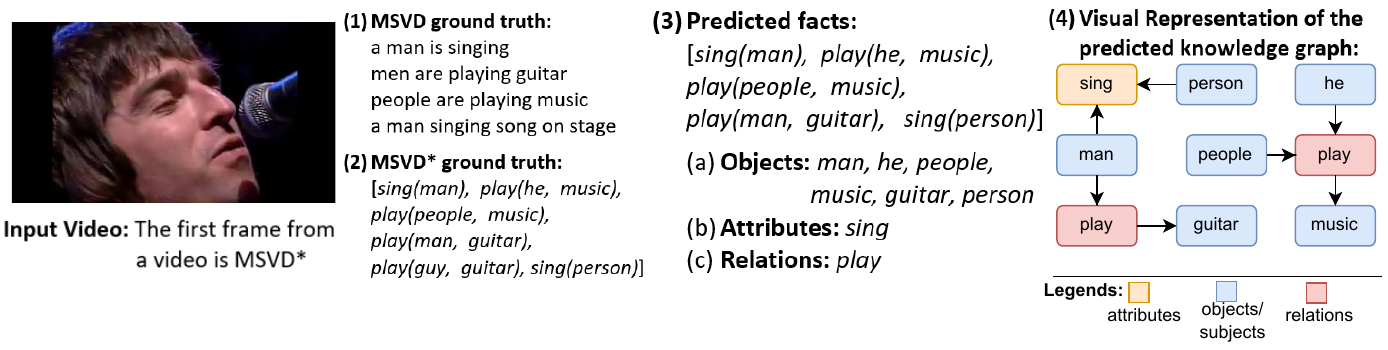}
    \vspace{-20pt}
    \caption{ \small The first frame from MSVD*, with (1) ground-truth natural language captions in MSVD, (2) the ground-truth set of facts in MSVD*, (3)the facts predicted by our model, with (a)objects/subjects present, (b)attributes predicted, (c)relations predicted, and (4) visual representation of the knowledge graph produced}
    \label{fig:example}
\end{figure*}

Let $X$ be the possible set of input videos. Let our vocabulary consist of a set $I$ of individuals and a set $P = C \cup R$ of predicates, where $C$ and $R$ are, respectively, unary and binary predicates. Our model then consists of 

\begin{enumerate}
    \item an encoder $f: X \rightarrow$ Z, where $Z$ is the space of extracted feature vectors, 
    \item three multi-classifiers, $g: Z \rightarrow (0,1)^{|I|}$, $h: Z \rightarrow (0,1)^{|C|}$ and $k: Z \rightarrow (0,1)^{|R|}$,
    \item A set of predicate multilayer perceptrons (predicate-MLPs), $\{m_p | p \in P\}$, 
    \item A set of trainable individual vectors: $\{v_i | i \in I \}$. 
\end{enumerate}

First, the input video $x$ is encoded using the encoder $f$ to produce the video encoding $e$. Then at the detection stage: the encoding is fed to the three multi-classifiers $g,h,k$, producing a prediction for each individual, each class, and each relation, respectively, being present in $x$; this corresponds to 3(a-c) in Figure \ref{fig:example}, and to the selection of nodes in the graph. Finally, in the fusion stage: we form a set of candidate unary and binary facts, for the given video, each of which is then evaluated directly by passing the individual vector(s) to the predicate-MLPs and thresholding the output. This corresponds to predicting the edges in the graph. 
%to produce the final annotation. This is an important step, as a poor selection of facts will degrade the model's performance (as seen in the ablation study in Section \ref{sec:ablation-studies}). 

Our vocabulary consists of $285$ individuals, $129$ attributes and $150$ relations. Hence, there are $258 \times 129 = 33,282$ candidate unary facts, and $ P_{2}^{258} \times 150 = 9,945,900$ candidate binary facts. Because running the MLPs is a computationally expensive step, we avoid running them on all combinations of individuals and predicates. Instead, we select only a subset of all combinations as candidate facts. 

Specifically, we note that the probability of a fact $c(s)$ occurring is upper-bounded by the joint probability of predicate $c$ and individual $s$ occurring, as it is a necessary condition that both $c$ and $s$ are in the video. For example, if the video shows a dog running, then the video must contain a dog and depict running. Note that this may not be sufficient, as there could be another object running while the dog is standing still. This observation gives a bound of $P(s \in I \cap c \in C | e)$, and similarly for binary facts (see supplementary material for the derivation of joint probability). The corresponding facts having the top $q$ values of joint probability are selected as the candidate facts and passed to the MLPs. The value of $q$ is a hyperparameter that we set to $1000$ in the main experiments. 
%We roughly choose the value experimentally as follows: each video has $\leq 3$ individuals in it: so, $3 \times 129$ pairs $+$ $P_{2}^{3} \times 150$ triples = $1287$ possible unary and binary facts. 

Section \ref{sec:ablation-studies} reports results for a wide range of values of $q$.

%Our vocabulary consists of $285$ individuals, $129$ attributes and $150$ relations (refer to supplementary material). Hence, there are $258 \times 129 = 33,282$ candidate unary facts, and $ P_{2}^{258} \times 150 = 9,945,900$ candidate binary facts. Typically each video has $\leq 3$ individuals in it. So, $3 \times 129$ pairs $+$ $P_{2}^{3} \times 150$ triples = $1287$ possible unary and binary facts. Hence, using this, we roughly choose a value of $q = 1000$ experimentally. A value of $< 1287$ is used to keep the overall inference time low. Section \ref{sec:ablation-studies} reports results for a wide range of values of $q$. 

%For each candidate fact, the final prediction is the output of the predicate MLP corresponding to the predicate in the fact, fed with the vectors for the predicted individual(s) in the fact along with encoding $e$. The probabilities received from the predicate MLPs are thresholded. The facts having probability values higher than the threshold are selected as true facts. The value of the threshold is decided experimentally, where the value between $0$ to $1$ giving the best F1-score on the training set of the dataset is selected.  

%Finally, all the true facts received are combined to produce the knowledge graph (KG) for the input video $x$. 

%\paragraph{Training}
To train the multiclassifier, we use the ground truth sets of individuals and predicates for each video, which are given explicitly in the datasets we use. That is, each multiclassifier is trained as in a standard multi-class, binary classification problem, using binary cross-entropy loss for each class. 

To train the predicate MLPs, we make use of the locally closed world assumption \citep{dong2014knowledge}, to avoid learning to predict everything as true. That is, the predicate MLPs are trained as in a standard binary classification problem, where the ground truth facts are the positive examples, and the facts that were corrupted using the locally closed-world assumption are the negative examples. These corrupted examples are also present explicitly in the datasets we use. 
All gradients are backpropagated to the encoder as well. 
%The loss for the encoder is 
%\begin{equation}
%     L = L_i + L_c + L_r + L_p \,,
%\end{equation}
%where $L_i$, $L_c$ and $L_r$ are the losses for the 3 multi classifiers, and $L_p$ is the loss from the predicate MLPs.

\paragraph{Inclusion of Background Knowledge}
%\label{sec:extend_model}
We propose a novel extension of our main model to use background/commonsense knowledge. For example, the model should favour predicting $drive(man, car)$ over $drink(man, car)$. Rather than try to explicitly encode intuitive physics that would express the impossibility of a man drinking a car, we instead use the statistics of how often given facts occur ``in nature''. There should be many occurrences of the $drive(man,car)$ and none of $drink(man,car)$, and this can be used to bias the model towards the former.

%As we will see in \ref{sec:ablation-studies}, this also increases the generalization capability of the model.

Specifically, in addition to the prediction from the main model, we produce another prediction based on statistics from an image dataset Visual Genome \citep{krishna2017visual}, which due to its large size and diversity, is here used to model the general prevalence of each fact. %Due to the lack of any such video dataset as per our knowledge, we are using an image dataset. 
%In addition to the prediction from the main model, we produce a second prediction for each candidate fact using this dataset. This is done as follows:
We extract the number of occurrences of each fact, of each subject and of each (subject,object) pair. Then, letting $A$ be the probability that a given fact is true, we can calculate the expectation of $A$, under the Bayesian posterior given the statistics in $VG$ as, $\mathbb{E}[A] = \int_{0}^{1} a \; P(a = A | D = d) \; da \,$, where $P(a = A | D = d)$  is the posterior probability of fact $a$ given its number of occurrence $d \in D$, in $VG$. Assuming a uniform prior, we can calculate the posterior as 

\begin{equation}\label{eq:10}
    \begin{split}
        P(a = A | D= d)  = \frac{P(D = d | a = A)  \times P(a = A)}{P(D =d)} = \frac{a^d \; (1-a)^{N-d} }{\frac{\Gamma(d + 1) \times \Gamma(-d + N + 1)}{\Gamma(N + 2)}} \,,\\
    \end{split}
\end{equation}
where $N$ is the total number of images and $\Gamma$ is the gamma function. This gives a final estimate of $\mathbb{E}[A] = \frac{d+1}{N+2}$ (see supplementary material for full derivation). 

\begin{comment}
    \begin{equation}
        \mathbb{E}[A] = \frac{d+1}{N+2}\,.
    \end{equation}
\end{comment}

\begin{figure*}
    \centering
    \includegraphics[width=0.8\textwidth]{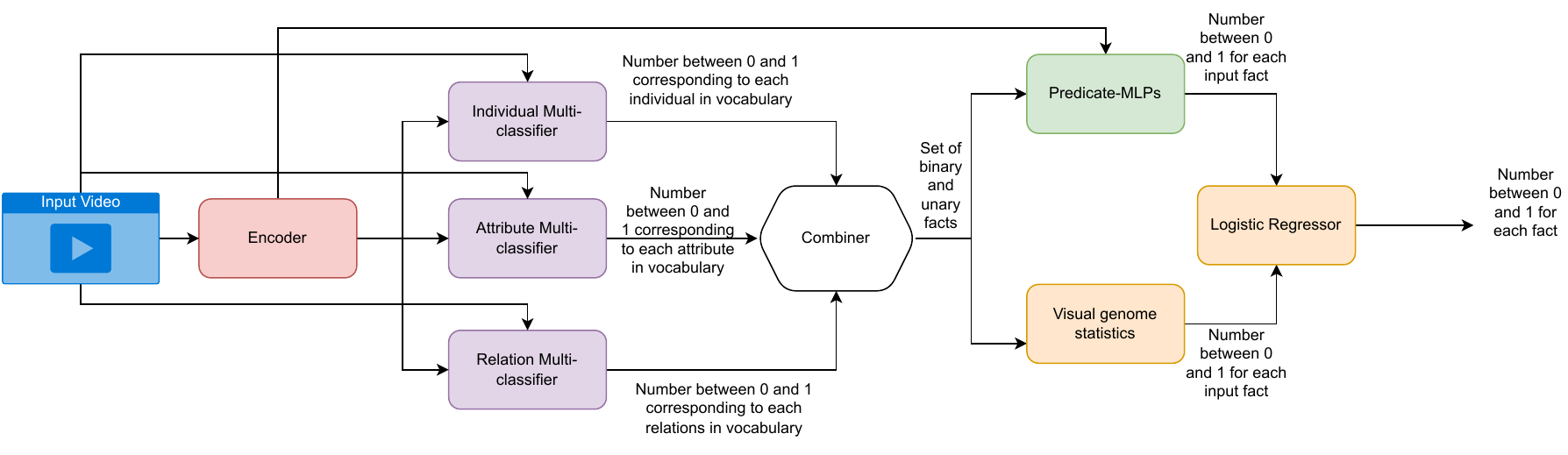}
    \vspace{-10pt}
    \caption{ \small Description of our approach for annotating a video input with knowledge graph using background knowledge as explained in Section\ref{sec:main_model}}
    \label{fig:test}
\end{figure*}

This estimate is then combined with that of the main model using a logistic regressor. The probabilities received from the logistic regressor are thresholded, to produce a prediction for all true facts. Finally, all the true facts received are combined to produce the knowledge graph for the input video $x$. Concisely, our proposed approach is shown in Figure \ref{fig:test}.
%The facts having probability values higher than the threshold are selected as true facts. 
%Finally, all the true facts received are combined to produce the knowledge graph for the input video $x$. Concisely, our proposed approach is shown in Figure \ref{fig:test}. 

%To train the logistic regressor, a separate dataset is built, where each training example has, as input, the probabilities produced by the main model and VG, and has the ground truth (either true or false) as the label. 

\paragraph{Implementation Details}
The encoder $f$ consists of a pre-trained VGG19 \citep{simonyan2014very} model followed by a 3-layer gated recurrent unit (GRU) \citep{cho2014learning}. As a second stream, we use a frozen copy of the I3D network \citep{Carreira_2017_CVPR}. We use these networks to allow comparison with \cite{mahon2020knowledge}. Results for other networks are reported in the supplementary material. The output of the encoder is a concatenation of this I3D feature vector and a weighted sum of the first stream, weighted by a learnable n-dimensional vector. The three multi-classifiers and each predicate's corresponding MLP have one hidden layer. The input size of the multi-classifiers is equivalent to the video encoding given by $dim(f(x)) = 5120$. While for predicate MLPs, the input size is $dim(f(x)) + D$, in case of unary facts and $dim(f(x)) + 2D$ in case of binary facts. Here $D$ is the size of individual(s) vectors ($300$ in our case). Weight updates are performed using Adam \citep{kingma2014adam}, with learning rate $.001$, $\beta_1=0.9$, $\beta_2=0.999$. Early stopping is employed during training with patience set to 7. %The value of $q$ in the main experiment is set to $1000$.
The logistic regressor and predicate MLP thresholds are selected as the values giving the best F1-score on the training set.

%Finally, we also select the value of $q$ from Section \ref{sec:main_model}. Our vocabulary consists of $285$ individuals, $129$ attributes and $150$ relations. Hence, there are $258 \times 129 = 33,282$ candidate unary facts, and $ P_{2}^{258} \times 150 = 9,945,900$ candidate binary facts. Typically each video has $\leq 3$ individuals in it. So, $3 \times 129$ pairs $+$ $P_{2}^{3} \times 150$ triples = $1287$ possible unary and binary facts. Hence, using this, we roughly choose a value of $q = 1000$ experimentally. A value of $< 1287$ is used to keep the overall inference time low. 

\section{Experimental Results} 
\label{sec:results}

%In this section, we first briefly describe the two datasets, MSVD* and MSRVTT*, used by our model. Then the quantitative experimental results on the datasets are reported and compared with other existing works. We also present some qualitative results, an ablation study and additional experiments. 

\subsection{Datasets}

We train and test our model on two automatically generated datasets for video annotation, taken from\cite{mahon2020knowledge}. 
The datasets MSVD* and MSRVTT* are generated from two well-known video captioning datasets: MSVD \cite{chen2011collecting} and MSRVTT \cite{xu2016msr}, respectively. Each training example contains captions in the form of a knowledge graph (KG), which is composed of a set of facts. Each fact contains a predicate and the corresponding arguments: $<subject, predicate, object>$ (binary facts), and $<subject, predicate>$ (unary facts). All the individuals and predicates are linked to entities in an ontology, WordNet \cite{miller1995wordnet}. 
%Words that appear $< 50$ times across the dataset and common semantically weak verbs are excluded from the dataset. For more information about the dataset and the generation please refer to\cite{mahon2020knowledge}.

\subsection{Main Results} \label{sec:main-reuslts}

The F1-score, positive, negative, and total accuracy scores generated using our model \ref{sec:main_model} for MSVD* and MSRVTT* datasets are given in Table \ref{tab:msvd_final_op}. The results are also compared with two existing works - (1) \cite{mahon2020knowledge} referred to as "LG 2020" here, and (2) \cite{vasile2018learning} referred to as "VL 2018". To the best of our knowledge, these two are the only existing works that have attempted the task of video annotation using KG and so are used to benchmark the performance of our system. 

As we can see, our system significantly outperforms both the models in F1-score, positive and total accuracy. Importantly, it gives superior positive accuracy, the most difficult metric to score highly on. The artificially constructed dataset we are using contains a higher percentage of negative facts than positive ones (refer\cite{mahon2020knowledge} for more information). This means that even if the model predicts everything as false, the negative accuracy would be very high. This issue has also been highlighted in \cite{mahon2020knowledge}, where the reported model was predicting most of the facts as negative. However, our system is not doing this and so the positive accuracy, as well as the F1 score, is far better. 

Interestingly, our results for MSRVTT* are significantly better than those for the MSVD*, even though, by most video captioning models in the literature, MSRVTT is considered a harder dataset \cite{xu2017learning, chen2018less, zhang2019reconstruct, zhang2017task, olivastri2019end}.

%Lastly, since the dataset is using an ontology, the WordNet \cite{miller1995wordnet}, inferring additional facts later on, is trivial. Though additional inferences are not produced in the present work, they can easily be done using WordNet. For example, the class $man$ is a subclass of $person$ and $male$. So, later on, it is possible to apply inference to all facts mentioning a $man$. If it is needed to determine how many videos in a database depict at least one person, or males, this would be possible here, but not in the case of merely annotating them with NL sentences. This is one of the advantages of our approach to using KGs for annotation. 

% \usepackage{graphicx}
\begin{table}[ht]
\centering
\caption{ \small F1, and positive/negative/total accuracy on MSVD* and the MSRVTT* datasets. The scores are average from 5 independent runs ($\pm$standard deviation). The best results are in bold.}
\label{tab:msvd_final_op}
\resizebox{1.0\columnwidth}{!}{%
\begin{tabular}{*{9}{c|}}
\cline{2-9}
& \multicolumn{4}{|c|}{MSVD*} & \multicolumn{4}{c|}{MSRVTT*}\\
\cline{2-9}
 & F1-score & \begin{tabular}[c]{@{}c@{}}Positive \\ Accuracy (\%)\end{tabular} & \begin{tabular}[c]{@{}c@{}}Negative \\ Accuracy (\%)\end{tabular} & \begin{tabular}[c]{@{}c@{}}Total \\ Accuracy\end{tabular} & F1-score & \begin{tabular}[c]{@{}c@{}}Positive \\ Accuracy (\%)\end{tabular} & \begin{tabular}[c]{@{}c@{}}Negative \\ Accuracy (\%)\end{tabular} & \begin{tabular}[c]{@{}c@{}}Total \\ Accuracy\end{tabular}\\ \hline
\multicolumn{1}{|c|}{\textbf{Ours}} & \textbf{27.13($\pm$1.42)} & \textbf{27.50($\pm$0.75)} & 89.99($\pm$0.73) & \textbf{79.90($\pm$1.17)} & \textbf{36.66($\pm$0.52)} & \textbf{36.36($\pm$0.58)} & 91.84($\pm$0.11) & 82.55($\pm$0.08)\\ \hline
\multicolumn{1}{|c|}{\textbf{LG 2020}} & 13.99 & 12.65 & \textbf{99.20} & 22.16 & 11.83 & 6.76 & \textbf{99.96} & \textbf{83.01} \\ \hline
\multicolumn{1}{|c|}{\textbf{VL 2018}} & 6.11 & 3.36 & - & - & - & - & - & - \\ \hline
\end{tabular}%
}
\end{table}

% \usepackage{graphicx}
%\begin{table}[ht]
%\centering
%\caption{ \small Results on the MSRVTT* video dataset annotation with KG, produced by the main model \ref{sec:main_model}. Best results in bold..}
%\label{tab:msrvtt_final_op}
%\resizebox{0.50\columnwidth}{!}{%
%\begin{tabular}{c|c|c|c|c|}
%\cline{2-5}
% & F1-score & \begin{tabular}[c]{@{}c@{}}Positive \\ Accuracy (\%)\end{tabular} & \begin{tabular}[c]{@{}c@{}}Negative \\ Accuracy (\%)\end{tabular} & \begin{tabular}[c]{@{}c@{}}Total \\ Accuracy\end{tabular} \\ \hline
%\multicolumn{1}{|c|}{\textbf{Ours}} & \textbf{37.23} & \textbf{34.07} & 94.15 & \textbf{84.08} \\ \hline
%\multicolumn{1}{|c|}{\textbf{LG 2020}} & 11.83 & 6.76 & \textbf{99.96} & 83.01 \\ \hline
%\end{tabular}%
%}
%\end{table}

\subsection{Inclusion of Background Knowledge}
\label{sec:result_extend_model}

%Here we present the quantitative results and further investigation of the results produced by the novel extension of our main model \ref{sec:extend_model}.

% \usepackage{graphicx}
\begin{table}[]
\centering
\caption{ \small Comparison between the main model and the extended model. The scores are average from 5 independent runs ($\pm$standard deviation). The best results are in bold.}
\label{tab:vg}
\resizebox{1.0\columnwidth}{!}{%
\begin{tabular}{*{9}{c|}}
\cline{2-9}
& \multicolumn{4}{|c|}{MSVD*} & \multicolumn{4}{c|}{MSRVTT*}\\
\cline{2-9}
%\resizebox{0.50\columnwidth}{!}{%
%\begin{tabular}{|l|l|l|l|l|}
%\hline
 &  F1-score &  \begin{tabular}[c]{@{}l@{}}Positive \\ Accuracy (\%)\end{tabular} &  \begin{tabular}[c]{@{}l@{}}Negative\\ Accuracy (\%)\end{tabular} &  \begin{tabular}[c]{@{}l@{}}Total\\ Accuracy\end{tabular}  &  F1-score &  \begin{tabular}[c]{@{}l@{}}Positive \\ Accuracy (\%)\end{tabular} &  \begin{tabular}[c]{@{}l@{}}Negative\\ Accuracy (\%)\end{tabular} &  \begin{tabular}[c]{@{}l@{}}Total\\ Accuracy\end{tabular} \\ \hline
\multicolumn{1}{|c|}{\textbf{Extended Model}} & \textbf{27.49($\pm$1.35)} & 27.46($\pm$1.23) & \textbf{90.62($\pm$1.4)} & \textbf{80.13($\pm$1.13)}  & 35.65($\pm$0.55) & \textbf{37.66($\pm$0.65)} & 89.81($\pm$0.14) & 81.08($\pm$0.12) \\ \hline
\multicolumn{1}{|c|}{\textbf{Main Model}} & 27.13($\pm$1.42) & \textbf{27.50($\pm$0.75)} & 89.99($\pm$0.73) & 79.90($\pm$1.17) & \textbf{36.66($\pm$0.52)} & 36.36($\pm$0.58) & \textbf{91.84($\pm$0.11)} & \textbf{82.55($\pm$0.08)} \\ \hline
\end{tabular}%
}
\end{table}

% \textbf{Extended} \\ \textbf{Model}
% \textbf{Main} \\ \textbf{Model}
% \usepackage{graphicx}
%\begin{table}[]
%\centering
%\caption{ \small Results on the MSRVTT* dataset, produced by the extended model \ref{sec:extend_model}. Best results in bold..}
%\label{tab:vg_msrvtt}
%\resizebox{0.50\columnwidth}{!}{%
%\begin{tabular}{|l|l|l|l|l|}
%\hline
% &
%  F1-score &
%  \begin{tabular}[c]{@{}l@{}}Positive \\ Accuracy (\%)\end{tabular} &
%  \begin{tabular}[c]{@{}l@{}}Negative\\ Accuracy (\%)\end{tabular} &
%  \begin{tabular}[c]{@{}l@{}}Total\\ Accuracy\end{tabular} \\ \hline
%\textbf{Extended Model} & 35.65 & 37.66 & 89.81 & 81.08 \\ \hline
%\textbf{Main Model} & \textbf{36.66} & 36.36 & \textbf{91.84} & \textbf{82.55} \\ \hline
%\end{tabular}%
%}
%\end{table}

%Table \ref{tab:vg_msvd} shows the effect of including background knowledge.
%, where each metric value is the average from 5 independent runs. 
As shown in Table \ref{tab:vg}, the inclusion of background knowledge produces a slightly better F1-score on MSVD* dataset, and positive accuracy on the MSRVTT* dataset. The reason we do not see a greater improvement may be because of the low overlap of components between our dataset and Visual Genome (see supplementary material for further details). 
%This in turn means many of the candidate facts or their components are not present in the Visual Genome dataset. For such cases, the final prediction in the extended model is only produced using the first set of predictions by predicate MLP, while ignoring predictions by $VG$. This decreases the overall effectiveness of using the Visual Genome predictions on the final output. However, even after this, there is a slight increase in the F1-score and positive accuracy for MSVD* and MSRVTT* dataset respectively, signifying that injecting commonsense knowledge is actually helping the model in avoiding the prediction of false facts. 

%
%Hence if future use of a better dataset, specifically made for videos, can further support the result generated by the proposed model. 
%
%
%
%\begin{table}[ht]
%\centering
%\resizebox{0.45\textwidth}{!}{%
%\begin{tabular}{|c|c|c|c|}
%\hline
%Dataset & 
%\begin{tabular}[c]{@{}c@{}}\% Overlap between \\ individuals\end{tabular} & \begin{tabular}[c]{@{}c@{}}\% Overlap between \\ attributes\end{tabular} & \begin{tabular}[c]{@{}c@{}}\% Overlap between \\ relations\end{tabular} \\ \hline
%\textbf{MSVD*} & 85.61 & 65.89 & 30.67 \\ \hline
%\textbf{MSRVTT*} & 88.77 & 63.76 & 33.33 \\ \hline
%\end{tabular}%
%}
%\caption{ \small Percentage of number of individuals, attributes and relations present in the Visual Genome dataset compared to MSVD* and MSRVTT*}
%\label{tab:VG_overlap}
%\end{table}
%

%\paragraph{Further Investigation of the contribution of Visual Genome statistics }
%\label{sec:ablation_VG}

\begin{figure}
    \centering
    \includegraphics[scale=0.24]{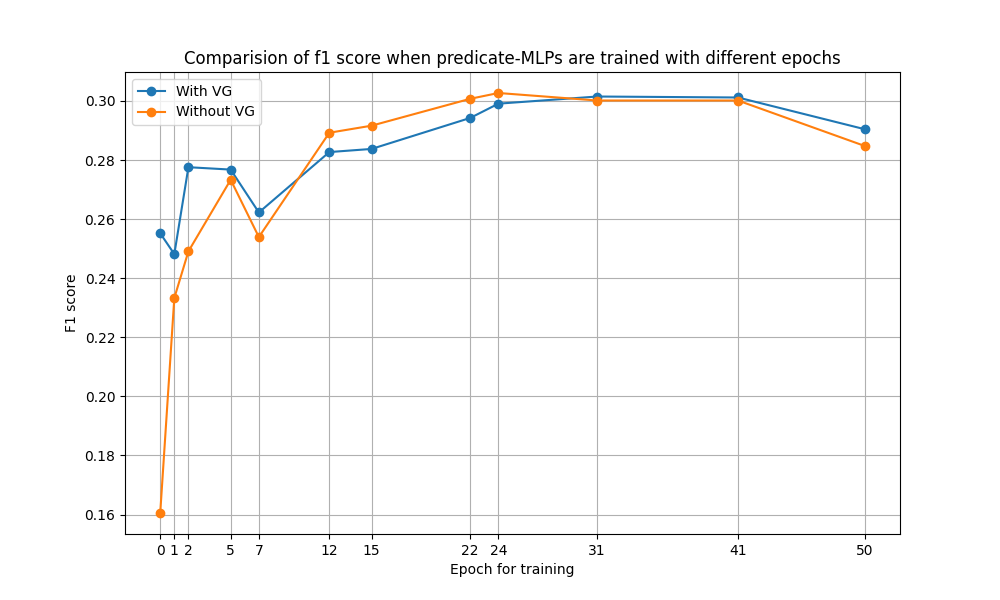}
    \includegraphics[scale=0.24]{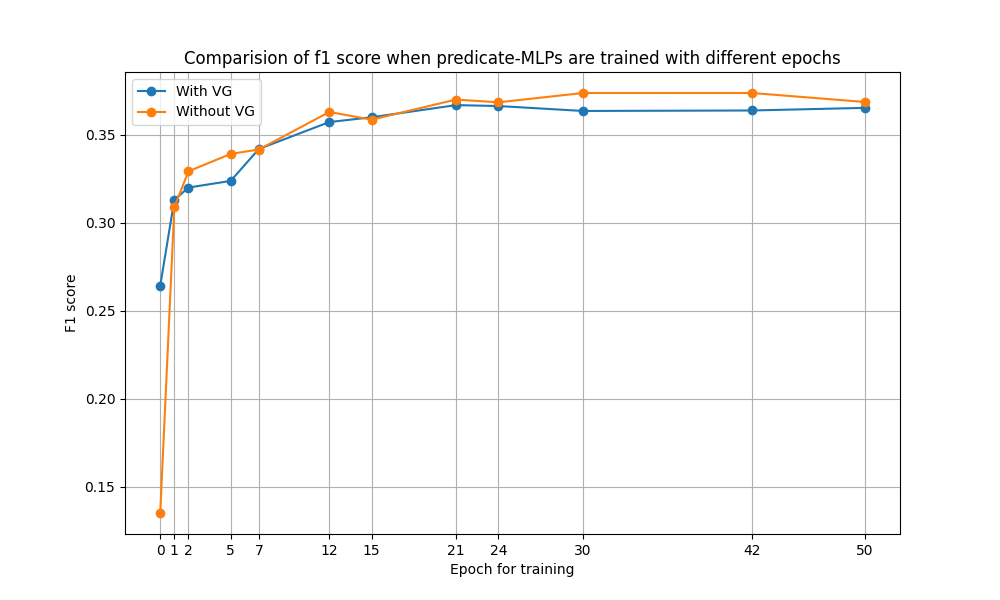}
        \vspace{-10pt}
    \caption{ \small F1-score vs the number of training epochs for the predicate-MLP, for the main model (left) and extended model (right).} 
    %The predicate-MLPs are trained at different epochs while the rest of the model is reloaded from a fully trained network. The plot shows the change in the F1-score with the increase in the number of training epochs for predicate-MLP training. 
    %At epoch=0 signifies that the predicate-MLP is untrained and is initialized with random weights.}
    \label{fig:ablation_epoch}
\end{figure}

To further understand how Visual Genome predictions are being used in an extended model, we examine the behaviour when the part of the model, the predicate MLPs component, is undertrained. Figure \ref{fig:ablation_epoch} shows F1-score when training of the predicate MLPs was stopped early. The x-axis shows the number of epochs the predicate MLPs were trained for. AT $x=0$, i.e. when the predicate-MLPs are untrained, the F1-score is far better when using the extended model. This signifies that when the network did not have any information about the dataset, $VG$ statistics representing general world knowledge helped the predictions the most. As we increase the number of epochs, the network learns more about the particular dataset, and the F1-score in both scenarios becomes close to each other, with the extended model ultimately giving a comparable result on the fully trained network.

%\begin{figure}
%    \centering
%    \includegraphics[scale=0.33]{images/MSRVTT_diffEpoch.png}
%    \caption{ \small Comparative results for MSRVTT* dataset, where "without VG" is the main model \ref{sec:main_model}, and "with VG" is the extended model \ref{sec:extend_model}. The predicate-MLPs are trained at different epochs while the rest of the model is reloaded from a fully trained network. The plot shows the change in the F1-score with the increase in the number of training epochs for predicate-MLP training. Epoch=0 signifies that the predicate-MLP is untrained and is initialized with random weights.}
%    \label{fig:msrvtt_ablation_epoch}
%\end{figure}

\subsection{Qualitative Results}
\label{sec:quali_result}

\begin{figure*}
    \centering
    \includegraphics[width=\textwidth]{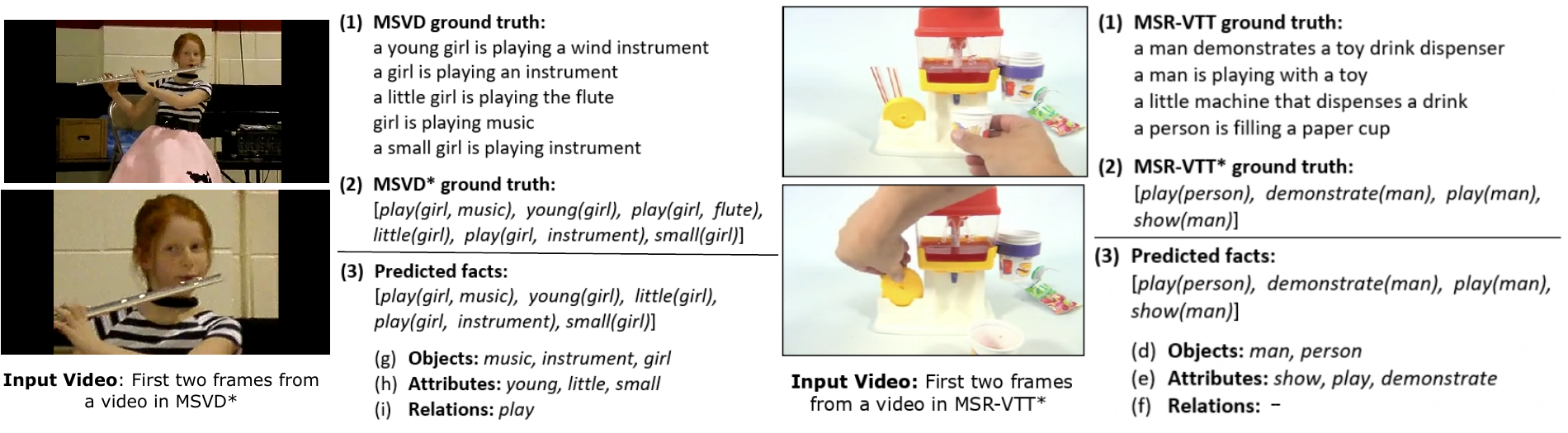}
    \vspace{-10pt}
    \caption{ \small Left: the first two frames from a video in MSVD*, with (1) ground-truth natural language captions in MSVD, (2) the ground-truth set of facts in MSVD*, (3) the facts predicted by the proposed model. Right: the first two frames from MSRVTT*, with (1) ground-truth natural language captions in MSRVTT, (2) the ground-truth set of facts in MSRVTT*, (3)the facts predicted by the proposed model.  Left: MSVD*, right: MSRVTT*.}
    \label{fig:msvd_quality}
\end{figure*}

To further evaluate the quality of the KGs produced for the video by our proposed model, manual inspection of videos and predicted facts is carried out. Figure \ref{fig:msvd_quality} shows the first two frames from a video with the facts predicted by the model for MSVD* (left) and MSRVTT* (right). These qualitative examples show the limitations imposed by the smaller vocabulary size in MSVD*. The individuals, attributes and relations which appear fewer than $50$ times are excluded from the dataset (see supplementary for further information on the dataset). 
%This implies that at times there could be insufficient material to describe a video fully. For example, the girl 
In the video in the left figure in Figure \ref{fig:msvd_quality}, the girl is playing the flute, which is also expressed in one of the MSVD captions. However, $flute$ appears less than 50 times and so is not in the model's vocabulary and it cannot predict $play(girl, flute)$. The model is, however, correctly able to identify it as an $instrument$, and also to correctly identify other attributes, relations, and individuals present in the video. 

The right of Figure \ref{fig:msvd_quality} shows the model correctly predicting all the facts for the video. However, it shows another limitation, which was beyond the scope of this work. In the synthetically generated dataset, the ground truth annotation does not express some facts about the video such as objects $paper\_cup$, $toy$, etc. They are excluded from the ground truth, so the model is not trained on them and may not predict them. This emphasizes the need for a manually generated structured video annotation dataset to avoid such cases.

\subsection{Ablation Studies} \label{sec:ablation-studies}

\paragraph{Ablation on combining framework}
%\label{sec:ablation_combiner}

As discussed in Section \ref{sec:main_model}, we produce candidate facts by combining the outputs from the individual-, attribute-, and relation-multi-classifiers. These candidate facts are later used to classify them as true or false to the given video and are an essential step in filtering irrelevant facts. To investigate the contribution of the combining procedure, we perform an ablation study on this network component. 

The results for MSVD* and MSRVTT* are shown in Table \ref{tab:msvd_ablation_comb}. 
In the "without combiner" setting, the combining framework is replaced by a simple threshold method, where the output of each multi-classifier is thresholded, and all permutations of the received individuals and predicates are used to build candidate facts. 
%That is,
%\begin{equation}
%    \begin{split}
%        I'   & = \{i \in I | g(f(x))_i > threshold\} \,,\\
%        C'   & = \{c \in C | h(f(x))_c > threshold\} \,,\\
%        R'   & = \{r \in R | k(f(x))_r > threshold\} \,,\\
%        U' & = I' \times C' \,,\\
%        B' & = I' \times I' \times R' \,,\\
%        T' & = U' \cup B' \,.
%        \label{eq:ablation_combiner}
%    \end{split}
%\end{equation}
The other components of the network are kept the same. 

The results in this setting are significantly worse than the main model, where the proposed combining framework is used. This is because many irrelevant facts are fed into the predicate-MLPs. This shows the effectiveness of the combining technique proposed in Section \ref{sec:main_model}.

% \usepackage{graphicx}
\begin{table}[ht]
\centering
\caption{ \small Ablation results on the combining technique given in Section \ref{sec:main_model}. The combining method here is replaced by a simple thresholding method. Best results in bold.}
\label{tab:msvd_ablation_comb}
\resizebox{1.00\columnwidth}{!}{%
\begin{tabular}{*{9}{c|}}
\cline{2-9}
& \multicolumn{4}{|c|}{MSVD*} & \multicolumn{4}{c|}{MSRVTT*}\\
\cline{2-9}
%\resizebox{0.50\columnwidth}{!}{%
%\begin{tabular}{c|c|c|c|c|}
%\cline{2-5}
 & F1-score & \begin{tabular}[c]{@{}c@{}}Positive \\ Accuracy (\%)\end{tabular} & \begin{tabular}[c]{@{}c@{}}Negative \\ Accuracy (\%)\end{tabular} & \begin{tabular}[c]{@{}c@{}}Total \\ Accuracy (\%)\end{tabular}  & F1-score & \begin{tabular}[c]{@{}c@{}}Positive \\ Accuracy (\%)\end{tabular} & \begin{tabular}[c]{@{}c@{}}Negative \\ Accuracy (\%)\end{tabular} & \begin{tabular}[c]{@{}c@{}}Total \\ Accuracy (\%)\end{tabular} \\ \hline
\multicolumn{1}{|c|}{\textbf{Main Model}} & \textbf{27.13} & \textbf{27.50} & 89.99 & 79.90 & \textbf{36.66} & \textbf{36.36} & 91.84 & 82.55 \\ \hline
\multicolumn{1}{|c|}{\textbf{Without Combiner}} & 10.6 & 7.75 & \textbf{98.96} & \textbf{83.65} & 13.6 & 9.01 & \textbf{99.66} & \textbf{84.47} \\ \hline
\end{tabular}%
}
\end{table}

% \usepackage{graphicx}
%\begin{table}[ht]
%\centering
%\caption{ \small Ablation results on the combining technique given in Section \ref{sec:main_model} on MSRVTT*. The combining method here is replaced by a simple thresholding method. Best results in bold.}
%\label{tab:msrvtt_ablation_comb}
%\resizebox{0.50\columnwidth}{!}{%
%\begin{tabular}{c|c|c|c|c|}
%\cline{2-5}
% & F1-score & \begin{tabular}[c]{@{}c@{}}Positive \\ Accuracy (\%)\end{tabular} & \begin{tabular}[c]{@{}c@{}}Negative \\ Accuracy (\%)\end{tabular} & \begin{tabular}[c]{@{}c@{}}Total \\ Accuracy (\%)\end{tabular} \\ \hline
%\multicolumn{1}{|c|}{Main Model} & \textbf{37.23} & \textbf{34.07} & 94.15 & 84.08 \\ \hline
%\multicolumn{1}{|c|}{Without Combiner} & 13.6 & 9.01 & \textbf{99.66} & \textbf{84.47} \\ \hline
%\end{tabular}%
%}
%\end{table}

\paragraph{Effect of the number of candidate facts}
%\label{sec:change_q}

As we discussed in Section \ref{sec:main_model}, after the detection stage, we choose the $q$ highest joint probability facts as candidate facts to pass to the fusion stage. The value of $q$ is a chosen hyperparameter and could be set to anything in the range $0 < q \leq 33282$ (as $|I|^2|R| = 33282$). Smaller values of $q$ would mean the number of candidate facts fed to predicate-MLPs is small, resulting in a smaller inference time, but the $F1$ score will be inferior. This is because many of the candidates could be false, as the dataset consists of more negative facts than positive facts. Feeding a bigger pool of facts to the predicate-MLPs increases the chances of receiving true facts but also increases inference time. 

\begin{figure}
    \centering
    \includegraphics[scale=0.15]{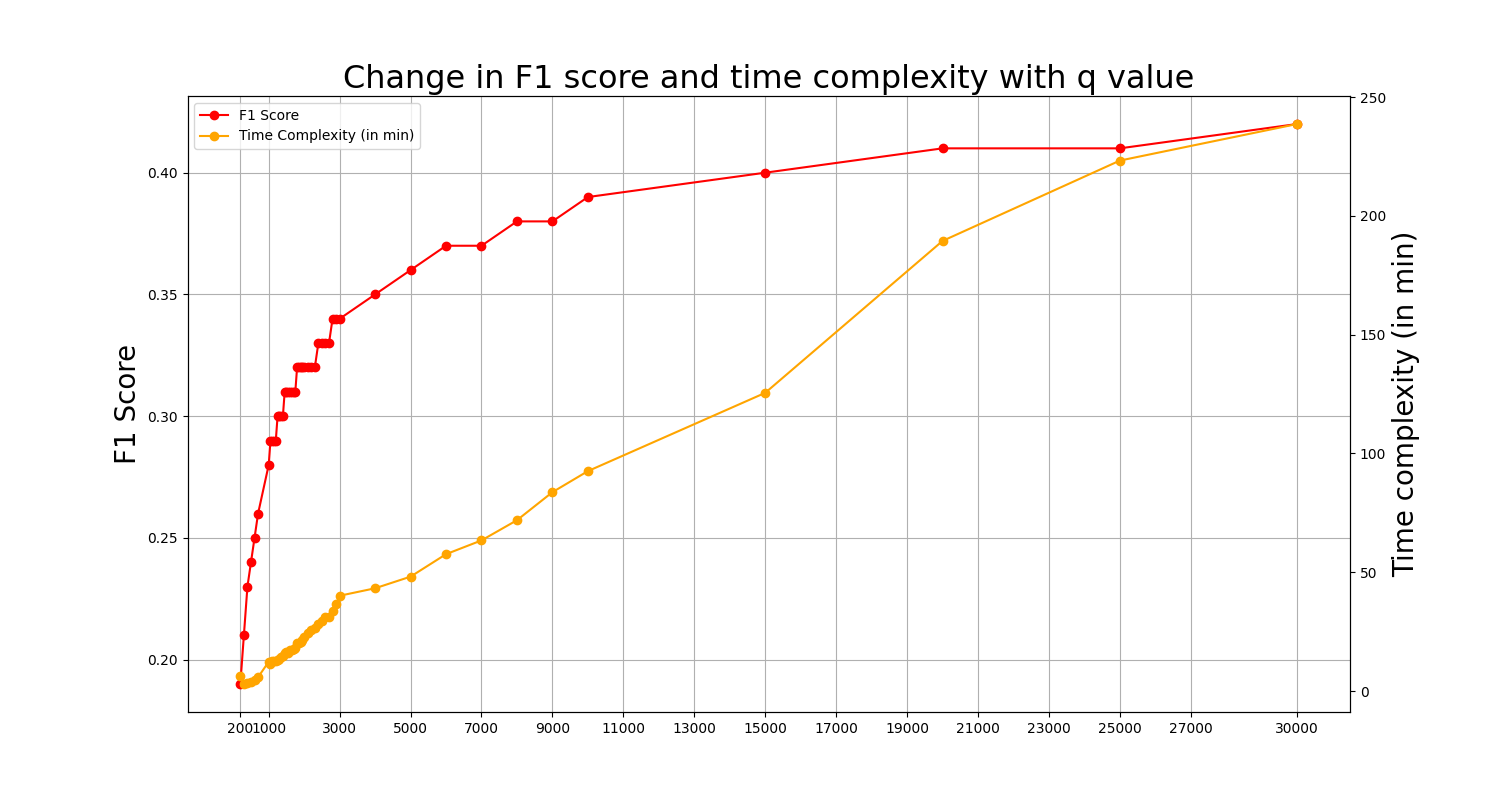}
    \includegraphics[scale=0.15]{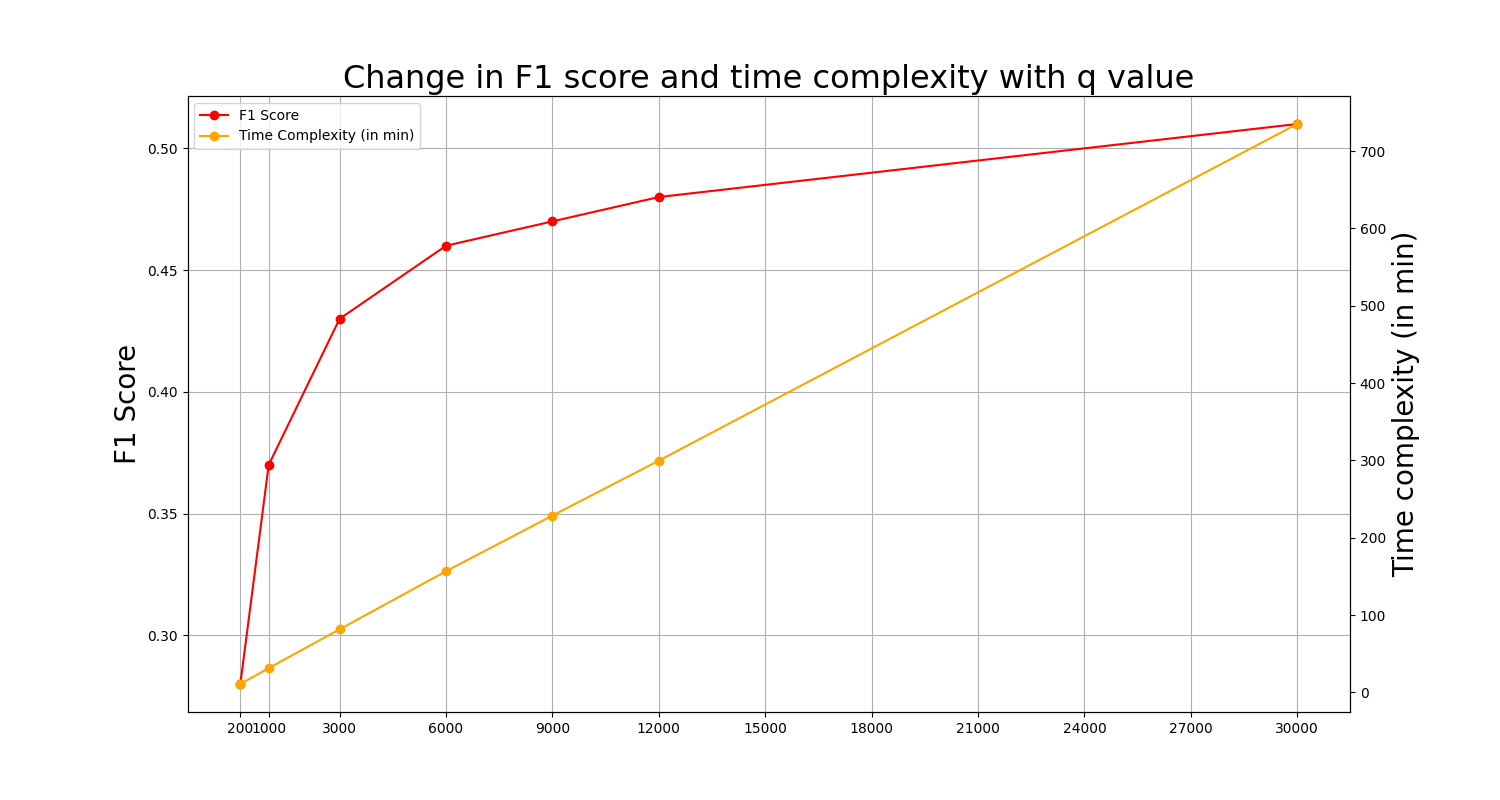}
    \vspace{-10pt}
    \caption{ \small Plots for changes in F1 and time-taken (in min) with changes in the number of candidate facts evaluated. Left: MSVD* dataset, right: MSRVTT* dataset.} 
    \label{fig:ablation_q}
\end{figure}

Here we perform experiments with different values of $q$ to study the effect of changing the value of $q$ on the overall performance of the main model. Figure \ref{fig:ablation_q} shows the performance (F1-score) as well as the time taken by the model to produce output, as $q$ is varied from $200 \leq q \leq 30000$. 
%Note that the time taken for inference depends on the processor used and the load on it at the time of the experiment, the time taken reported here is only representative of the actual value. Since the experiments are conducted on shared GPU servers, the load on them has varied throughout the experiment, which affects the overall experiment time. 
%Figure \ref{fig:ablation_q} shows the plots for MSVD* and MSRVTT* datasets respectively. 
The 'red curve' plots the F1-score on the y-axis with the corresponding $q$ value on the x-axis. The 'yellow line' on the plot shows the time taken for the inference (min) on the second y-axis, corresponding to the $q$ value on the x-axis and the F1-score on the first y-axis. 

Interestingly, we can see that the F1-score continuously grows, with an almost exponential increase for the first few epochs, while for higher epochs, the growth rate slows down, with a nearly linear increase in time taken for inference, with an increase in value of $q$. As expected, the F1-score grows fast initially, with decreasing growth rate with every increase in $q$. This implies that with better computational resources, our model will be able to achieve ever better performance by using a higher value of $q$. %This is superior to the other work reported in section \ref{sec:main-reuslts}. 

\section{Conclusion} \label{sec:conclusion}

This paper proposed a new deep-learning model for the task of KG extraction from videos. The KG here is composed of a set of facts that describes the relations held between individuals. We also explore the inclusion of background knowledge in the construction of KG. %As per our knowledge, our work is the first to explore the inclusion of background knowledge in KG. 
%We provide arguments and empirical support for the advantage of background knowledge in KG. 
Further, we evaluate both our main and extended models, both qualitatively and qualitatively, and present extensive investigative and ablation studies showing the contribution of various components of our model. Our model significantly outperforms existing models and has much better generalization capability.

Future works include exploring the use of other datasets for injecting commonsense which is more exhaustive and comprehensive. It is also interesting to explore KG extraction from other input domains. Such as application to text, where the model could perform a task similar to open information extraction. Another extension could be to manually construct a dataset designed for video annotation using KG, rather than relying on automatically generated datasets.

\bibliography{egbib}
\end{document}

\maketitle

\section{Derivation of Joint Probability}

As we see in Section 3.1, the probability of a fact $c(s)$ occurring is upper-bounded by the joint probability of $c$ and
$s$ occurring, as it is a necessary condition that both $c$ and
$s$ are in the video. We present the derivation of the joint probability here. 

The multi-classifiers $g$, $h$ and $k$ produce predictions for each individual, each class, and each relation, respectively, being present in input video $x$, when fed with the video encoding $e$. Hence, the joint probability is given by:

\begin{equation}\label{eq:comb_classif1}
     = 
    \begin{cases}
        P((s \in I \cap c \in C) | e), \; \text{if unary fact}  \\
       P((s \in I \cap o \in I \cap r \in R) | e), \; \text{if binary fact} \,,\\
    \end{cases}
\end{equation}

where $I$ is the set of all individuals in the vocabulary, $C$ is the set of all unary predicates in the vocabulary, $R$ is the set of all binary predicates in the vocabulary.
%and $a$ is the given fact.

Since, $g$, $h$ and $k$ are conditionally independent given the input video encoding $e$, we get, 
\begin{equation}
    \begin{split}
    \label{eq:joint_prob}
        P((s \in I \cap c \in C) | e) & = P(s \in I | e) \times P(c \in C | e), \; \text{if unary fact} \,,\\
        P((s \in I \cap o \in I \cap r \in R) | e) & = P(s \in I \cap o \in I| e) \times P(r \in R | e), \; \text{if binary fact} \,.
    \end{split}
\end{equation}

Equation \ref{eq:joint_prob} for binary facts can be further simplified. Let the hidden layer in $g$ is denoted by $l$ (the multi-classifier contains only 1 hidden layer). Then we can assume that, $P(s \in I)$ and $P(o \in I)$ are conditionally independent given the $e$ and $l$. That is, 

\begin{equation}\label{eq:s_o_indep}
    P(s \in I \cap o \in I| e) = P(s \in I | l, e) \times P(o \in I | l, e), \;\; s \indep o |l, e \,.
\end{equation}

Finally, using Equation \ref{eq:s_o_indep}, the joint probability is given by:

\begin{equation}\label{eq:comb_classif2}
     = 
    \begin{cases}
        P(s \in I | e) \times P(c \in C | e), \; \text{if unary fact} \,. \\
      P(s \in I | e, l) \times P(o \in I | e, l) \times P(r \in R | e), \; \text{if binary fact} \,. \\
    \end{cases}
\end{equation}

Equation \ref{eq:comb_classif2} simply means that the joint probability is given by the multiplication of the probability of each component in the fact. Hence, for producing candidate facts, the respective probabilities produced by the multi-classifiers are multiplied, followed by the selection of top $q$ values.

\section{Derivation of Second Prediction using Visual Genome}

In Section 3.1 (Inclusion of Background Knowledge), we propose the use of the Visual Genome dataset to inject background/commonsense knowledge by producing a second prediction for each fact using this dataset. The full derivation of it is presented here. 

Let $A$ be the probability that a given fact is true. Then,
the second prediction for each fact is given by the expected value of $A$ under the Bayesian posterior $P(A = a | D = d)$ given the statistics $VG$. 

\begin{equation}\label{eq:4}
    \mathbb{E}[A] = \int_{0}^{1} a \; P(A = a | D = d) \; da \,,
\end{equation}

where $P(A = a | D = d)$  is the posterior probability, $a$ is the possible probability value that a given fact is true, and $D = d$ is the number of occurrences of the fact in $VG$. The prediction produced by the Visual Genome dataset for a fact is given by $\mathbb{E}[A]$.

Using Bayes' theorem \cite{swinburne2004bayes}, the posterior probability of the given fact can be written as,  
\begin{equation}\label{eq:5}
    P(A = a | D = d) = \frac{P(D = d | A = a) \times P(A = a)}{P(D = d)} \,,
 \end{equation}
where $P(A = a)$ is the Bayesian prior, $P(D = d | A = a)$ is the likelihood and $P(D = d)$ is the evidence. 

For simplicity, let us assume a uniform prior, i.e. $P(A = a) = 1$, such that \\ $\int_{0}^{1} \; P(A = a) \; da = 1$. 

\paragraph{Calculation of likelihood probability:}

Now, the likelihood is given by:
\begin{equation}\label{eq:9}
    \begin{split}
        P(D =d | A = a) = a^d \; (1-a)^{N-d} \,,
    \end{split}
\end{equation}
where $N$ is given by Equation \ref{eq:7}.

\paragraph{Calculation of evidence:}

\begin{equation}
    \begin{split}
        \label{eq:6}
            P(D = d) & = \int_{0}^{1} x^d (1-x)^{N-d} \,dx \,,
    \end{split}
\end{equation}

where $N$ is the total number of images in the Visual Genome dataset containing the $subject$ (or $(subject, object)$ pair for a binary fact) for the given fact.
That is, 
\begin{equation}\label{eq:7}
    N = 
    \begin{cases}
        \#(\_,s) \text{if}\,, len(a) = 2 \\
        \#(\_,s,o) \text{if}\,, len(a) = 3 \,,\\
    \end{cases}
\end{equation}
where $s$ is the subject in unary fact, and $(s,o)$ is the subject-object pair in binary fact.
%Why is P(D = d) given by this? 
%Maximum Aposteriori (MAP) distribution of possible values. 

Coming back to equation \ref{eq:6},

\begin{align*}\label{eq:pd}
    %\begin{split}
        P(D = d) & = \int_{0}^{1} x^d (1-x)^{N-d} \,dx \numberthis \\
        & \text{Using integration by parts, i.e., $\int_{a}^{b} u \frac{dv}{dx}dx = uv - \int v \frac{du}{dx}dx$} \,,\\
        & \text{Substituting $u = (1-x)^{N-d}$ and $\frac{dv}{dx} = x^d \Rightarrow v = \frac{x^{d+1}}{d+1} + constant$} \,, \displaybreak\\
                & = (\frac{x^{d+1} (1-x)^{N-d}}{d+1} + constant) \Big|_0^1 + \frac{N-d}{d+1} \int_{0}^{1} x^{d+1} (1-x)^{N-d-1} dx \;, \\    & \text{for} \; Re(d - N)<1 \wedge Re(d)>-2 \\
                & \text{As $d \in \mathbb{Z}_{\geq 0}$, $d \leq N$, $Re(d - N)<1 \wedge Re(d)>-2$ is always satisfied} \,,\\
                & = 0 + \frac{N-d}{d+1} \int_{0}^{1} x^{d+1} (1-x)^{N-d-1} dx \\
                & = \frac{N-d}{d+1} \cdot \frac{N-d-1}{d+2} \cdot\int_{0}^{1} x^{d+2} (1-x)^{N-d-2} dx \\
                & \text{Repeating integrating by parts several times, we get} \,,\\
                & = \frac{(N-d) \cdot (N-d-1) \cdot (N-d-2) \dots 2 \cdot 1}{(d+1) \cdot (d+2) \dots N} \int_{0}^{1} x^N (1-x)^{0} dx \\
                & = \frac{d! (N-d)!}{N!} \int_{0}^{1} x^N dx \\
                & = \frac{d! (N-d)!}{N!} \cdot \frac{1}{N+1} \,,\\
                & = \frac{d! (N-d)!}{(N + 1)!} \\
                & \text{Using Fact 1, we get} \,,\\
                & = \frac{\Gamma(d + 1) \times \Gamma(-d + N + 1)}{\Gamma(N + 2) } \,,
    %\end{split}
\end{align*}

where $\Gamma$ is the gamma function, $Re(z)$ is the real part of $z$, $e_1 \wedge e_2 \wedge \dots$ is the logical AND function, and $\mathbb{Z}_{\geq 0} = \{0,1,2 \dots\}$ denotes the set of whole numbers.

\paragraph{Calculation of posterior probability:}
Finally, using equations \ref{eq:5}-\ref{eq:pd}, the posterior probability is given by:

\begin{equation}\label{eq:10}
    \begin{split}
        P(A = a | D= d) & = \frac{P(D = d | A = a)  \times P(A = a)}{P(D =d)} \,,\\
                        & = \frac{a^d \; (1-a)^{N-d}  \times 1}{\frac{\Gamma(d + 1) \times \Gamma(-d + N + 1)}{\Gamma(N + 2) }} \,,\\
                        & = \frac{a^d \; (1-a)^{N-d} }{\frac{\Gamma(d + 1) \times \Gamma(-d + N + 1)}{\Gamma(N + 2)}} \,,\\
                        & = \frac{a^d \; (1-a)^{N-d} }{\frac{d! \times (-d + N)!}{(N + 1)!}} \,.\\
    \end{split}
\end{equation}

Finally, using the equations \ref{eq:4}-\ref{eq:10}, the expected value for a given fact is calculated as, 

\begin{align*}\label{eq:11}
    %\begin{split}
        \mathbb{E}[A] & = \int_{0}^{1} a \; P(A = a | D = d) \; da \numberthis \\
                    & = \int_{0}^{1} \; \frac{a \times (a^d \; (1-a)^{N-d})}{\frac{\Gamma(d + 1) \times \Gamma(-d + N + 1)}{\Gamma(N + 2)}} \; da \\
                      & = \frac{1}{\frac{\Gamma(d + 1) \times \Gamma(-d + N + 1)}{\Gamma(N + 2)}} \times \int_{0}^{1} \; a \times (a^d \; (1-a)^{N-d}) \; da \\
                      & = \frac{\Gamma(N + 2)}{\Gamma(d + 1) \times \Gamma(-d + N + 1)} \times \int_{0}^{1} \; a^{d+1} \; (1-a)^{N-d}) \; da \\
                      & \text{Substituting solution of integral $\int_{0}^{1} \; a^{d+1} \; (1-a)^{N-d})$ using Equation \ref{eq:substitute_int}, we get,}\\
                      & = \frac{\Gamma(N + 2)}{\Gamma(d + 1) \times \Gamma(-d + N + 1)} \times \frac{\Gamma(d + 2) \times \Gamma(-d + N + 1)}{\Gamma(N + 3)} \\
                      \text{Using Fact 1} \,,\\
                      & = \frac{\Gamma(d + 2)}{\Gamma(d + 1)} \times \frac{\Gamma(N + 2)}{\Gamma(N + 3)} \\
                      & = \frac{(d+1)!}{d!} \times \frac{(N+1)!}{N+2!} \\
                      & = \frac{d+1}{N+2} \,.
    %\end{split}
\end{align*}

\paragraph{Fact 1.} 
\label{fact:1}
$\Gamma$ function is a generalized factorial function, i.e.  $\Gamma(n+1) = n!$, for all non-negative whole numbers $n$. Since $d$ and $N$ are the counts they are always non-negative whole numbers, this condition applies in our case.

\begin{align*}\label{eq:substitute_int}
    %\begin{split}
        Int & = \int_{0}^{1} a^{d+1} (1-a)^{N-d} \,dx \,, \numberthis \\
        & \text{Using integration by parts, given by $\int_{a}^{b} u \frac{dv}{dx}dx = uv - \int v \frac{du}{dx}dx$} \,,\\
        & \text{Substituting $u = a^{d+1}$ and $\frac{dv}{da} = (1-a)^{N-d} \Rightarrow v = \frac{(1-a)^{N-d+1}}{d-N-1} + constant$} \,, \\
                & = (\frac{a^{d+1} (1-a)^{N-d+1}}{d-N-1} + constant) \Big|_0^1 + \frac{d+1}{N-d+1} \int_{0}^{1} a^{d} (1-a)^{N-d+1} da \;, \\ &\text{for} \; Re(d - N)<1 \wedge Re(d)>-2 \,,\\
                & \text{As $d \in \mathbb{Z}_{\geq 0}$, $d \leq N$, $Re(d - N)<1 \wedge Re(d)>-2$ is always satisfied} \,,\\
                & = 0 + \frac{d+1}{N-d+1} \int_{0}^{1} a^{d} (1-a)^{N-d+1} da \,, \\
                & = \frac{d+1}{N-d+1} \cdot \frac{d}{N-d+2} \cdot\int_{0}^{1} a^{d-1} (1-a)^{N-d+2} da \,,\\
                & \text{Repeating integrating by parts several times, we get} \,, \\
                & = \frac{(d+1) \cdot (d) \cdot (d-1) \dots 2 \cdot 1}{(N-d+1) \cdot (N-d+2) \dots (N+1)} \int_{0}^{1} a^0 (1-a)^{N+1} da \,, \\
                & = \frac{(d+1)! (N-d)!}{(N+1)!} \int_{0}^{1} a^{N+1} dx \,,\\
                & = \frac{(d+1)! (N-d)!}{(N+1)!} \cdot \frac{1}{N+2} \,,\\
                & = \frac{(d+1)! (N-d)!}{(N + 2)!} \,,\\
                & \text{Using Fact 1, we get} \,,\\
                & = \frac{\Gamma(d + 2) \times \Gamma(-d + N + 1)}{\Gamma(N + 3) } \,.
    %\end{split}
\end{align*}

\section{Comparison with Different Network}

As discussed in Section 3.1 (Implementation Details), the pre-trained VGG19 \citep{simonyan2014very} is one of the networks employed in the encoder. 
Here we present and compare results from another network, namely Resnet152 \cite{he2016deep}. Hence, here the encoder $f$ consists of a pre-trained Resnet152 model, followed by a 3-layer gated recurrent unit (GRU) \citep{cho2014learning}. As a second stream, we use a frozen copy of the I3D network \citep{Carreira_2017_CVPR}. The rest of the components of the model is kept the same as given in Section 3.

Table \ref{tab:resnet152} compares the results produced by the main model with the Resnet152 network and VGG19 network for MSVD* and MSRVTT* datasets. Resnet152 gives a marginal improvement for MSVD*, and no improvement for MSRVTT*, which suggests that the feature extraction network is not the bottleneck of the entire model.

\begin{table}[ht]
\centering
\caption{ \small Comparison between main model with Resnet152 network and VGG19 network. F1, and positive/negative/total accuracy reported on MSVD* and the MSRVTT* datasets. The scores are averaged from 8 independent runs ($\pm$standard deviation). The best results are in bold.}
\label{tab:resnet152}
\resizebox{1.0\columnwidth}{!}{%
\begin{tabular}{*{9}{c|}}
\cline{2-9}
& \multicolumn{4}{|c|}{MSVD*} & \multicolumn{4}{c|}{MSRVTT*}\\
\cline{2-9}
 & F1-score & \begin{tabular}[c]{@{}c@{}}Positive \\ Accuracy (\%)\end{tabular} & \begin{tabular}[c]{@{}c@{}}Negative \\ Accuracy (\%)\end{tabular} & \begin{tabular}[c]{@{}c@{}}Total \\ Accuracy\end{tabular} & F1-score & \begin{tabular}[c]{@{}c@{}}Positive \\ Accuracy (\%)\end{tabular} & \begin{tabular}[c]{@{}c@{}}Negative \\ Accuracy (\%)\end{tabular} & \begin{tabular}[c]{@{}c@{}}Total \\ Accuracy\end{tabular}\\ \hline
\multicolumn{1}{|c|}{\begin{tabular}[c]{@{}c@{}}Main Model with \\ Resnet152 Network\end{tabular}} &
\textbf{27.91($\pm$1.94)} & \textbf{27.65($\pm$1.78)} & \textbf{90.92($\pm$1.09)} & \textbf{80.31($\pm$0.96)} & 36.59($\pm$0.7) & 36.34($\pm$0.83) & 91.76($\pm$0.21) & 82.48($\pm$0.1)\\ \hline
\multicolumn{1}{|c|}{\begin{tabular}[c]{@{}c@{}}Main Model with \\ VGG19 Network\end{tabular}} &
27.13($\pm$1.42) & 27.50($\pm$0.75) & 89.99($\pm$0.73) & 79.90($\pm$1.17) & \textbf{36.66($\pm$0.52)} & \textbf{36.36($\pm$0.58)} & \textbf{91.84($\pm$0.11)} & \textbf{82.55($\pm$0.08)} \\ \hline
\end{tabular}%
}
\end{table}

Similarly, Table \ref{tab:resnet152_extended} compares the results produced by the extended model using Restnet152 and VGG19 network. Resnet152 gives a marginal improvement for MSRVTT*, and on some metrics (F1 score, positive and total accuracy) for MSVD* dataset.

\begin{table}[ht]
\centering
\caption{ \small Comparison between extended model with Resnet152 network and VGG19 network. F1, and positive/negative/total accuracy reported on MSVD* and the MSRVTT* datasets. The scores are averaged from 5 independent runs ($\pm$standard deviation). The best results are in bold.}
\label{tab:resnet152_extended}
\resizebox{1.0\columnwidth}{!}{%
\begin{tabular}{*{9}{c|}}
\cline{2-9}
& \multicolumn{4}{|c|}{MSVD*} & \multicolumn{4}{c|}{MSRVTT*}\\
\cline{2-9}
 & F1-score & \begin{tabular}[c]{@{}c@{}}Positive \\ Accuracy (\%)\end{tabular} & \begin{tabular}[c]{@{}c@{}}Negative \\ Accuracy (\%)\end{tabular} & \begin{tabular}[c]{@{}c@{}}Total \\ Accuracy\end{tabular} & F1-score & \begin{tabular}[c]{@{}c@{}}Positive \\ Accuracy (\%)\end{tabular} & \begin{tabular}[c]{@{}c@{}}Negative \\ Accuracy (\%)\end{tabular} & \begin{tabular}[c]{@{}c@{}}Total \\ Accuracy\end{tabular}\\ \hline
\multicolumn{1}{|c|}{\begin{tabular}[c]{@{}c@{}}Extended Model with \\ Resnet152 Network\end{tabular}} &
\textbf{27.75($\pm$2.27)} & 27.45($\pm$1.74) & \textbf{91.04($\pm$1.23)} & \textbf{80.37($\pm$1.16)} & \textbf{36.25($\pm$0.82)} & \textbf{37.88($\pm$0.78)} & \textbf{90.14($\pm$0.2)} & \textbf{81.39($\pm$0.17)}\\ \hline
\multicolumn{1}{|c|}{\begin{tabular}[c]{@{}c@{}}Extended Model with \\ VGG19 Network\end{tabular}} &
27.49($\pm$1.35) & \textbf{27.46($\pm$1.23)} & 90.62($\pm$1.4) & 80.13($\pm$1.13)  & 35.65($\pm$0.55) & 37.66($\pm$0.65) & 89.81($\pm$0.14) & 81.08($\pm$0.12) \\ \hline
\end{tabular}%
}
\end{table}

\section{Datasets}

Our model is trained and tested on two automatically generated datasets referred to as MSVD* and MSRVTT* here. The dataset is taken from \cite{KG_extraction_dataset}. Here we present a brief overview of the generation of these two datasets. 

Two well-known video-captioning datasets, annotated with natural language (NL) captions, MSVD \cite{chen2011collecting} and MSRVTT \cite{xu2016msr}, are taken and processed to produce MSVD* and MSRVTT*, respectively. 

First, the NL sentences in the caption are parsed using a rule-based parser based on the Stanford NLP syntactic parser \cite{qi2019universal}. This produces a dependency parse of the sentence, where the part of speech for each word and the syntactic relations that hold true of them are identified. Then, a sequence of rules is applied to the above syntactic parse to form atoms. The atoms are then used to form facts, where each fact contains a predicate and the corresponding arguments. Next, word-sense disambiguation is performed by linking all the predicates and subjects to ontology. Word-sense disambiguation is necessary for polysemous words, which are words with more than 1 distinct sense. The ontology used for linking is WordNet \cite{miller1990introduction}. Now, the set of atoms (linked to WordNet) forms the corresponding logical annotation. All the logical annotations for each video are merged by taking the union of all the constituent atoms. After this, semantically weak verbs and all the words that appear less than 50 times across the dataset are excluded. This is because words appearing a few times would not produce a good performance. Verbs such as "do", "take", "be", and "have" are considered semantically weak verbs. They generally function as copulas or syntactic operators and do not convey relevant information about the video. The above method produces the knowledge graph for each video. Lastly, the above method only produces non-negated facts. This would imply that the model trained on this data could simply learn to predict every potential fact to be true. Hence, to solve this issue, the negated facts are also produced using local closed-world assumption (LCWA) \cite{dong2014knowledge}. For detailed information about the dataset and dataset generation steps please refer to \cite{KG_extraction_dataset, mahon2020knowledge}

\subsection{Overlap between Our Datasets and Visual Genome}

In Section 4.3, we discuss the extended model with the Visual Genome dataset. Table \ref{tab:VG_overlap} shows the overlap of components in our datasets and Visual Genome. As we can see there is a low overlap between our datasets and the Visual genome, especially for attribute and relation complements. This decreases the overall effectiveness of using the Visual Genome predictions on the final output and could be a reason why we do not see greater improvement in the results produced by the extended model. 

\begin{table}[ht]
\centering
\caption{Percentage of number of individuals, attributes and relations present in the Visual Genome dataset compared to MSVD* and MSRVTT* datasets.}
\label{tab:VG_overlap}
\resizebox{.8\textwidth}{!}{%
\begin{tabular}{|c|c|c|c|}
\hline
Dataset & 
\begin{tabular}[c]{@{}c@{}}\% Overlap between \\ individuals\end{tabular} & \begin{tabular}[c]{@{}c@{}}\% Overlap between \\ attributes\end{tabular} & \begin{tabular}[c]{@{}c@{}}\% Overlap between \\ relations\end{tabular} \\ \hline
\textbf{MSVD*} & 85.61 & 65.89 & 30.67 \\ \hline
\textbf{MSR-VTT*} & 88.77 & 63.76 & 33.33 \\ \hline
\end{tabular}%
}
\end{table}

\bibliography{egbib}